\documentclass[conference,a4paper,10pt]{IEEEtran}
\IEEEoverridecommandlockouts
\usepackage{texnames}
\usepackage{bm,bbm}
\usepackage{pbalance}
\usepackage[cmex10]{amsmath}
\usepackage{amssymb,amsfonts}
\usepackage{algorithmic}
\usepackage{algorithm}
\usepackage{array}
\usepackage[final]{changes}

\usepackage{caption, setspace}
\usepackage{subcaption}

\usepackage{textcomp}
\usepackage{stfloats}
\usepackage{verbatim}
\usepackage{graphicx}
\usepackage[normalem]{ulem}

\usepackage{amssymb}
\usepackage{tabularx}
\usepackage{multirow}
\usepackage{siunitx}
\usepackage{pifont}
\usepackage{graphicx}
\usepackage{amsfonts}
\usepackage{booktabs}
\usepackage{blindtext}
\usepackage{csquotes}
\usepackage{makecell}
\usepackage{tikz}
\usepackage{tikzscale}
\usepackage{float}
\usepackage[above]{placeins}

\usepackage[style=ieee, doi=true, url=true, isbn=false, minnames=1, maxnames=10, date=year, maxcitenames=2, hyperref=true]{biblatex}
\defbibheading{secbib}[\bibname]{%
  \section{#1}%
  \markboth{#1}{#1}
}
\AtBeginBibliography{\footnotesize}
\renewbibmacro{in:}{}
\DeclareBibliographyDriver{unpublished}{%
    \usebibmacro{bibindex}%
    \usebibmacro{begentry}%
    \usebibmacro{author/translator+others}%
    \newunit
    \usebibmacro{title}%
    \newunit\newblock
    \usebibmacro{eprint}%
    \newunit\newblock
    \usebibmacro{date}
    \usebibmacro{finentry}%
}

\DeclareFieldFormat{eprint:arxiv}{%
  arXiv\addcolon\space
  \ifhyperref
    {\href{https://arxiv.org/abs/#1}{
       \nolinkurl{#1}%
       \iffieldundef{eprintclass}
         {}
         }}
    {
    #1
     \iffieldundef{eprintclass}
       {}
       }
}
\DeclareSIUnit\epochs{epochs}
\usepackage[acronym]{glossaries}
\usepackage[printonlyused]{acronym}
\usepackage{etoolbox}
\usepackage{enumitem}    
\newlist{compactenum}{enumerate}{4}
\setlist[compactenum,1]{nosep,label*=\arabic*)}

\usepackage[shortcuts]{extdash}
\usepackage{comment}
\usepackage[normalem]{ulem}
\newcommand{\hl}[1]{#1}
\usepackage[hidelinks]{hyperref}
\usepackage[capitalise,noabbrev]{cleveref} 
\crefname{section}{section}{sections}  
\Crefname{Section}{Section}{Sections}
\usepackage{orcidlink}

\Crefname{figure}{Fig.}{Figs.}

\newcommand{\numpatches}{\num{549488}}
\newcommand{\zenodourl}{\url{https://doi.org/10.5281/zenodo.10891137}}

\newcommand{\BEN}{BigEarthNet}
\newcommand{\BENnew}{reBEN}

\newacronym{msi}{MSI}{multispectral imaging}
\newacronym{esa}{ESA}{European Space Agency}
\newacronym{dl}{DL}{deep learning}
\newacronym{rs}{RS}{remote sensing}
\newacronym{sar}{SAR}{synthetic aperture radar}
\newacronym{cnn}{CNN}{convolutional neural network}
\newacronym{vit}{ViT}{vision transformer}
\newacronym{lidar}{LIDAR}{light detection and ranging}
\newacronym{hsi}{HSI}{hyperspectral imaging}
\newacronym{mml}{MML}{multi-modal~learning}
\newacronym{lstm}{LSTM}{long short-term memory}
\newacronym{lulc}{LULC}{land use land cover}
\newacronym{clc}{CLC}{CORINE Land Cover}
\newacronym{grn}{GRN}{Global Response Normalization}

\begin{document}





\title{\scshape \BENnew{}: Refined BigEarthNet Dataset for Remote Sensing Image Analysis}

\author{
    \IEEEauthorblockN{Kai Norman Clasen$^{*\,1,2}$, %
    Leonard Hackel$^{*\,1,2}$, %
    Tom Burgert$^{1,2}$, %
    Gencer Sumbul$^{3}$,
    Begüm Demir$^{1,2}$, %
    Volker Markl$^{1,2,4}$}\\
    \IEEEauthorblockA{
    \hfill
    $^1$\,TU Berlin, Germany \hfill $^2$\,BIFOLD, Germany \hfill
    $^3$\,EPFL, Switzerland \hfill
    $^4$\,DFKI, Germany
    \hfill}
    \thanks{This work is supported by the European Research Council
(ERC) through the ERC-2017-STG BigEarth Project under
Grant 759764 and 
by the European Space Agency through the DA4DTE (Demonstrator precursor Digital Assistant interface for Digital Twin Earth) project.}

}

\maketitle
\def\thefootnote{*}\footnotetext{These authors contributed equally to this work.}\def\thefootnote{\arabic{footnote}}

\begin{abstract}
This paper presents refined BigEarthNet (reBEN) that is a large-scale, multi-modal remote sensing dataset constructed to support deep learning (DL) studies for remote sensing image analysis. The reBEN dataset consists of 549,488 pairs of Sentinel-1 and Sentinel-2 image patches.
To construct reBEN, we initially consider the Sentinel-1 and Sentinel-2 tiles used to construct the BigEarthNet dataset and then divide them into patches of size 1200\,m $\times$ 1200\,m.
We apply atmospheric correction to the Sentinel-2 patches using the latest version of the sen2cor tool, resulting in higher-quality patches compared to those present in BigEarthNet.
Each patch is then associated with a pixel-level reference map and scene-level multi-labels.
This makes \BENnew{} suitable for pixel- and scene-based learning tasks.
The labels are derived from the most recent CORINE Land Cover (CLC) map of 2018 by utilizing the 19-class nomenclature as in BigEarthNet.
The use of the most recent CLC map results in overcoming the label noise present in BigEarthNet. 
Furthermore, we introduce a new geographical-based split assignment algorithm that significantly reduces the spatial correlation among the train, validation, and test sets with respect to those present in BigEarthNet.
This increases the reliability of the evaluation of DL models. 
To minimize the DL model training time, we introduce a software tool (called as rico-hdl) that converts the reBEN dataset into a DL-optimized data format. In our experiments, we show the potential of reBEN for multi-modal multi-label image classification problems by considering several state-of-the-art DL models.
The pre-trained model weights, associated code, and complete dataset are available at \url{https://bigearth.net}.
\end{abstract}

\begin{IEEEkeywords}
Multi-modal learning, multi-label image classification, deep learning, remote sensing.
\end{IEEEkeywords}

\section{Introduction}\label{sec:introduction}
The rapid evolution of satellite systems results in a significant growth of \gls{rs} image archives.
For an accurate analysis of the vast amount of data available, it is necessary to define automatic methods.
To this end, developing \gls{dl} based methods is a growing research interest in the \gls{rs} community \cite{ai-to-advance-eo}.
To support the development and assessment of such methods, several large-scale benchmark datasets have been presented in \gls{rs} \cite{ben_S2,BigEarthNetMM,ai4smallfarms}.
A comprehensive list of \gls{rs} benchmark datasets is given in \cite{rsdatasetsurvey}.
Among the existing datasets, one of the widely used datasets is \BEN{}  \cite{ben_S2, BigEarthNetMM} (also known as \BEN{}-MM).
\BEN{} is a large-scale multi-modal and multi-label benchmark dataset presented in the framework of remote sensing image classification and retrieval problems.
It consists of Sentinel-1 and Sentinel-2 image patches covering \replaced{ten}{10} European countries.
With its release, it has paved the way for the design and development of
different benchmark datasets in \gls{rs}.
For instance, in \cite{benge} the ben-ge dataset is constructed by adding geographical (e.g., climate-zone and topographic data) and environmental (e.g., weather and seasonal information) data to the BigEarthNet patches.
As another example, the RSVQAxBEN dataset comprises BigEarthNet patches, as well as semantic content-related questions and answers formulated in natural language, which is presented in \cite{rsvqaxben} for visual question answering (VQA) problems. 

Although \BEN{} has been proven as an effective dataset in \gls{rs},
we have recently identified some issues:
\begin{compactenum}
\item Atmospheric correction tool updates:
The Sentinel-2 tiles of the BigEarthNet dataset have been pre-processed with the atmospheric correction tool  sen2cor \cite{sen2cor} version 2.5.5 \cite{sen2corV255} from the \gls{esa}, as it was the recommended version during the dataset creation.
Since then, updates to the sen2cor pre-processor have improved the output of processed Sentinel-2 tiles.
Due to the different outputs of the atmospheric correction tools, \gls{dl} models trained on BigEarthNet may not perform as well on tiles pre-processed with recent versions \cite{sen2corV211}.
\item  \Gls{lulc} label noise:
During the construction of the BigEarthNet dataset, the land use and land cover classes were derived from the preliminary \gls{clc} map of 2018 (CLC2018) \cite{clcbook}.
However, since then the CLC2018 map has received several updates that address wrong and missing label annotations (label noise).
Thus, the BigEarthNet dataset contains label noise that negatively affects the training and evaluation of \gls{dl} models \cite{dortmund}.
Furthermore, the BigEarthNet dataset does not include pixel-level reference maps, making it unsuitable for pixel-based learning tasks.
\item Training, validation, and test split correlation:
The recommended split in the \BEN{} dataset has a high spatial correlation among the training, validation and test sets, making it difficult to evaluate results reliably.
\item Limited software tools for efficient and effective \gls{dl} model training:
Loading and processing the \BEN{} image data 
takes a considerable amount of time, which is especially noticeable when training \gls{dl} models.
\item Lack of recent pre-trained models: While pre-trained \gls{dl} models were made available with the release of the BigEarthNet dataset, several \gls{dl} architecture advancements have been developed in the meantime. Such state-of-the-art models pre-trained on \BEN{} are currently not available.
\end{compactenum}

All of these issues affect the development and evaluation of BigEarthNet-based datasets and \gls{dl} models using these datasets.
To overcome this problem, we introduce the \BENnew{} dataset\footnote{\zenodourl} that addresses the above-mentioned issues, leading to more reliable
and interpretable research in \gls{rs} image analysis compared to the BigEarthNet dataset.

\section{\BENnew{} Dataset}\label{sec:dataset-description} 
This section provides a detailed description of the \BENnew{} dataset.

\subsection{\BENnew{} patch generation and labeling procedure}\label{subsec:patch-generation}

\begin{figure}[tb!]
    \centering
    \setlength{\fboxrule}{.2pt}
    \setlength{\fboxsep}{0pt}
    \def\subfigfrac{.3}
    \hfill
    \begin{subfigure}[t]{\subfigfrac\linewidth}
        \fbox{\includegraphics[width=\linewidth]{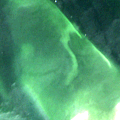}}\caption{
            \BEN{}: {Marine waters} (M), Urban fabric (W) \\
            \BENnew{}: Marine waters (C)
        }\label{subfigure:wrong_label:11_33}
    \end{subfigure} \hfill
    \begin{subfigure}[t]{\subfigfrac\linewidth}
        \fbox{\includegraphics[width=\linewidth]{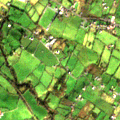}}\caption{
            \BEN{}: {Pastures (C)}, {Urban fabric (M)} \\
            \BENnew{}: {Pastures (C)}, {Urban fabric (C)}        }\label{subfigure:wrong_label:12_38}
    \end{subfigure} \hfill\hfill
    \caption{\label{fig:wrong_label} Two example patches with the associated multi-labels from the BigEarthNet and \BENnew{} datasets, 
    where the correct labels are indicated by (C), wrong labels by (W) and missing labels by (M).
    }
\end{figure}

To generate \BENnew{} patches, we initially downloaded the same 125 Sentinel-2 level-1C tiles (which are associated to less than \SI{1}{\percent} cloud cover acquired between June 2017 and May 2018) considered in \BEN{} from the new Copernicus Data Space Ecosystem (CDSE).
These tiles were selected to represent each geographical location in different seasons.
To obtain the atmospherically corrected and ortho-rectified bottom-of-atmosphere reflectance data product Sentinel-2 level-2A, we utilized the most recent version of the atmospheric correction tool sen2cor (version 2.11) \cite{sen2corV211}.

After obtaining the 125 Sentinel-2 level-2A tiles, the metadata embedded within each level-2A tile was inspected. The metadata contains quality indicator fields (e.g., for radiometric and geometric values) that inform the user about potential issues related to the respective tile. %
For a detailed description of all quality indicator values and what checks are associated with them, the reader is referred to the official Sentinel-2 product specification document \cite{sentinel2spec}. 
Among the 125 tiles, 6 tiles fail radiometric and geometric quality indicator checks. 
We have decided not to include these tiles to ensure high-quality Sentinel-2 data in \BENnew{}. Then, the remaining tiles were split into patches covering a region of \SI{1200}{\metre} $\times$ \SI{1200}{\metre} each. 
Note that the resulting patches cover the exact same geographical areas as the patches in the \BEN{} dataset.
\replaced{In the \BENnew{} dataset, patches containing pixels with invalid data are not considered. However, those that are fully covered by seasonal snow, clouds and cloud shadows are kept as in \BEN{}, and are listed in a dedicated file.
The associated Sentinel-1 patches from \BEN{} are incorporated into the dataset to produce the complete multi-modal \BENnew{} dataset.}{In the \BENnew{} dataset, patches that contain pixels with invalid data are not considered, whereas those that are fully covered by seasonal snow, clouds and cloud shadows are listed in a separate file but kept, as in the case of \BEN{}.
The Sentinel-1 patches from BigEarthNet are added to \BENnew{} to produce the complete multi-modal \BENnew{}.}

\begin{figure}[tb!]
    \centering
    \setlength{\fboxrule}{.2pt}
    \setlength{\fboxsep}{0pt}
    \def\subfigfrac{.3}
    \hfill
    \begin{subfigure}{\subfigfrac\linewidth}
        \fbox{\includegraphics[width=\linewidth]{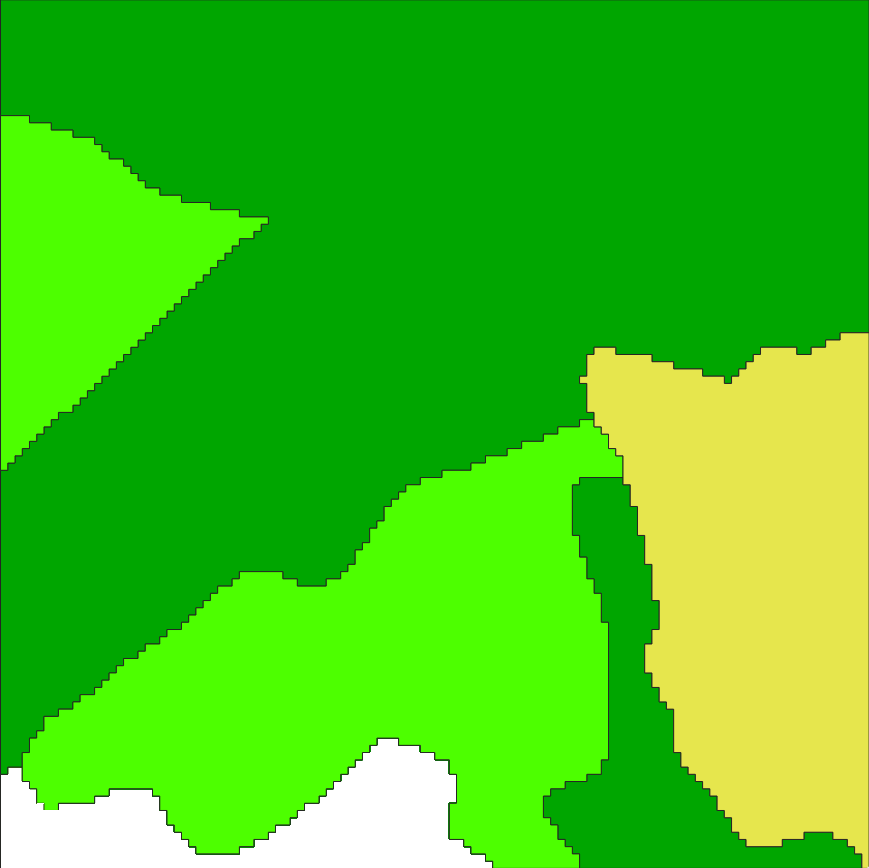}}\caption{}\label{subfigure:patch:partial}
    \end{subfigure} \hfill
    \begin{subfigure}{\subfigfrac\linewidth}
        \fbox{\includegraphics[width=\linewidth]{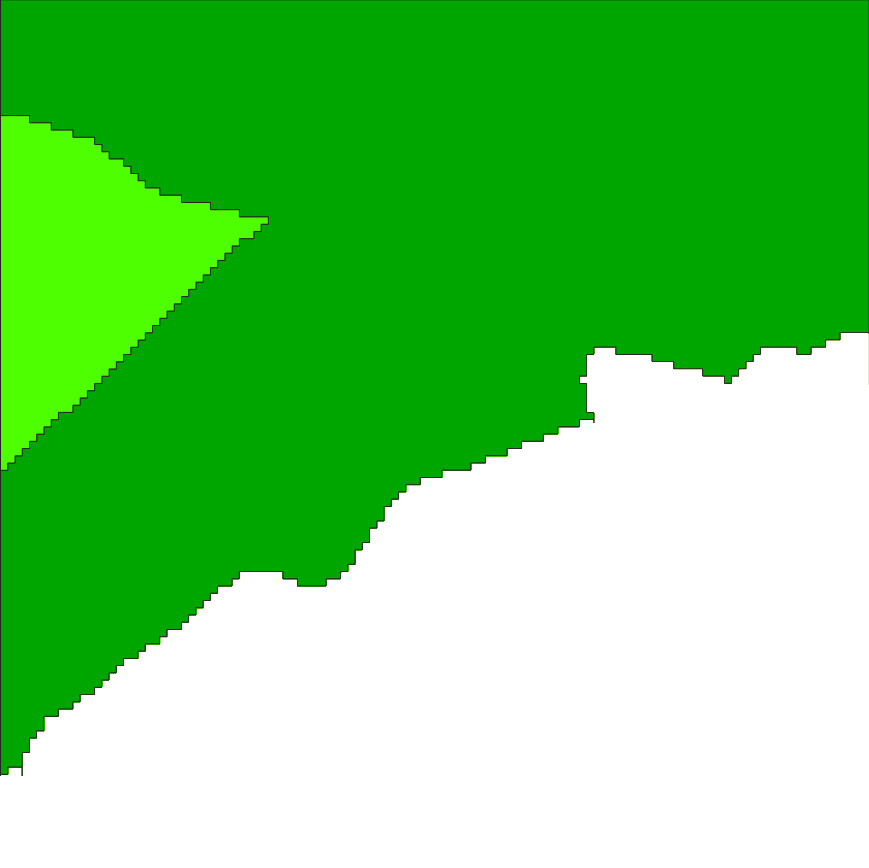}}\caption{}\label{subfigure:patch:partialdropped}
    \end{subfigure}
    \hfill\hfill
    \caption{\label{fig:patchlbl}
    An example of pixel-level reference maps associated to different amounts of unlabeled pixels depicted in white: 
    (a) shows a reference map that has a small number of unlabeled pixels whose associated patch is not removed;
    and (b) shows a reference map that has less than \SI{75}{\percent} of pixels annotated, thus its associated patch is excluded from the \BENnew{} dataset.}
\end{figure}

After constructing the patches, each patch is associated with a pixel-level reference map
obtained by overlaying the considered patch with the polygons from the most recent CLC2018 map (\texttt{V2020\_20u1}) \cite{clc2018}.
This allows the use of the \BENnew{} dataset for pixel-based learning tasks.
Additionally, multi-labels are extracted from the pixel-level reference maps to exploit \BENnew{} for scene-based learning tasks (e.g., multi-label scene classification or image retrieval).
Due to the use of the most recent CLC2018 \cite{clc2018} label map, the \gls{lulc} label noise problem present in BigEarthNet is addressed (see \cref{fig:wrong_label}).
From \cref{subfigure:wrong_label:11_33}, one can see that a patch with the label \enquote{Marine waters} was mislabeled as \enquote{Urban fabric} in BigEarthNet, while it is correctly labeled in the \BENnew{} dataset.
\Cref{subfigure:wrong_label:12_38} shows a patch where BigEarthNet only contains the \enquote{Pastures} label, however, it is missing the \enquote{Urban fabric} label.
In \BENnew{}, the patch is correctly annotated with both labels. Note that the \gls{lulc} class labels of \BENnew{} are defined based on the 19-class nomenclature as in \cite{BigEarthNetMM}.

Some of the \BENnew{} patches are either only partially labeled or completely unlabeled. This is due to two main reasons. One reason is that the most recent CLC2018 map \cite{clc2018} covers a smaller geographical area than the one used in BigEarthNet.
The other reason is due to the use of the 19-class nomenclature.
We have decided to exclude the patches with no labels from the \BENnew{} dataset.
However, removing all patches that are not fully annotated by class labels would be overly stringent.
Therefore, we only remove a patch if less than \SI{75}{\percent} of its pixels are annotated. 
The value of \SI{75}{\percent} was identified based on a validation procedure that provides a balanced tradeoff between potentially associated label noise due to missing label information and excluding an excessive number of patches.
\Cref{fig:patchlbl} shows an example of pixel-level reference maps.
The map given in \Cref{subfigure:patch:partial} has only a small number of unlabeled pixels. Thus, its associated patch is not excluded. 
However, the patch associated with the map shown in \cref{subfigure:patch:partialdropped} is excluded as less than \SI{75}{\percent} of the pixels are labeled.
Due to the above-mentioned reasons, the total number of the remaining pairs of patches in \BENnew{} 
is \numpatches{}, which is slightly smaller than that of BigEarthNet.
We would like to note that we make the code\footnotemark{} to construct \BENnew{} publicly available to support reproducibility. This can also allow one to easily extend the dataset to other desired geographical areas.

\footnotetext{\url{https://github.com/rsim-tu-berlin/bigearthnet-pipeline}}

\subsection{Recommended training, validation, and test split}\label{subsec:new-split}

\begin{figure}[tb!]
    \centering
    \setlength{\fboxrule}{.2pt}
    \setlength{\fboxsep}{0pt}
    \def\subfigfrac{.3}
    \hfill
    \begin{subfigure}{\subfigfrac\linewidth}
        \fbox{\includegraphics[width=\linewidth]{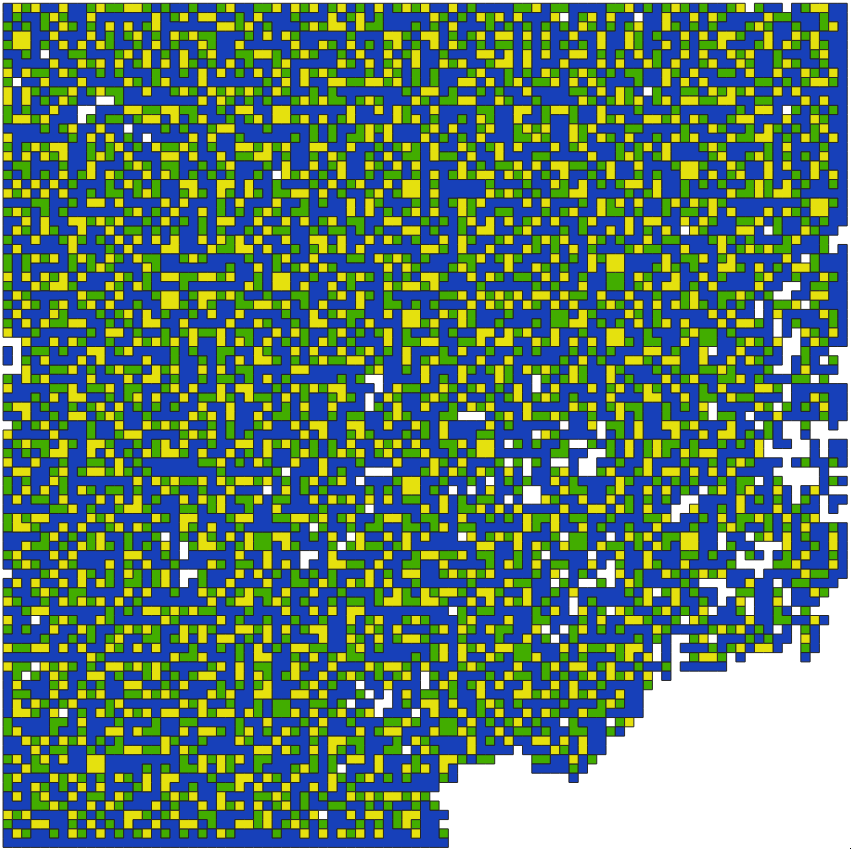}}\caption{}\label{subfigure:split:old}
    \end{subfigure} \hfill
    \begin{subfigure}{\subfigfrac\linewidth}
        \fbox{\includegraphics[width=\linewidth]{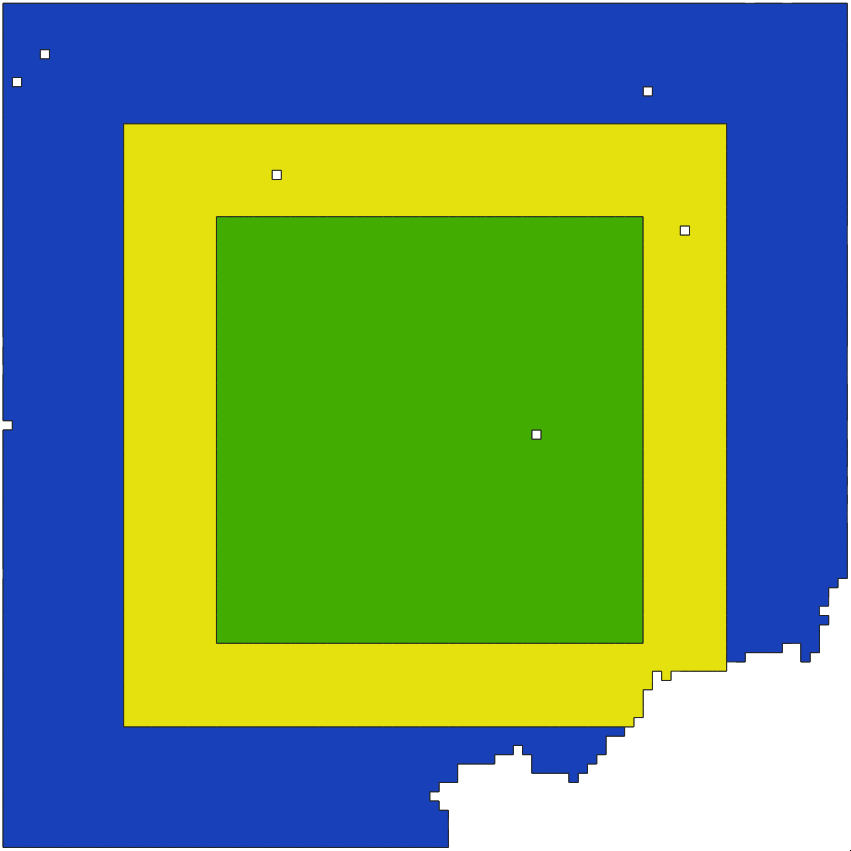}}\caption{}\label{subfigure:split:new}
    \end{subfigure}
    \hfill\hfill
    \caption{\label{fig:split}
    \hl{Results of the BigEarthNet and \BENnew{} split assignment algorithms on one of the 119 tiles.}
    The patches of the training, validation and test sets are colored in blue, yellow, and green, respectively.
    The uncolored areas represent invalid patches.
    (a) shows the results obtained by the grid-based split assignment algorithm from BigEarthNet with patches from different sets within close spatial proximity to each other; and
    (b) shows the results obtained by the geographical-based split assignment algorithm of \BENnew{} with larger distances between patches of different splits.
    }
\end{figure}

The training, validation, and test split in BigEarthNet is constructed by using a grid-based split assignment algorithm.
\Cref{subfigure:split:old} visualizes the spatial distribution of the sets for an example tile.
From the figure, one can observe that this algorithm results in highly correlated sets, which prevents reliable evaluation of the results.

To address this issue, we introduce a new geographical-based split assignment algorithm.
Our algorithm divides the area of a tile into an inner square surrounded by an inner and outer frame as shown in \cref{subfigure:split:new}.
We assign the patches in the outer frame to the training set, those in the inner frame to the validation set, and the remaining ones to the test set (see \Cref{subfigure:split:new}).
This results in a distinct separation among the patches of the training, validation, and test sets, while maximizing the distance between the training and test sets.
It is important to note that adjacent Sentinel-2 tiles have overlapping borders. 
Assigning the training set to the outer frame ensures that the overlapping patches acquired within the same season are located within the same split.
By this way, these \hl{overlapping} patches do not implicitly skew the evaluation results.
This is an important enrichment compared to the split used in \BEN{}, where overlapping patches are assigned to separate sets.

We aim to produce a similar number of patches per set as in the BigEarthNet dataset.
Given the total area of a Sentinel-2 tile, which is guaranteed to be a square with size $s \times s$, the width of the outer frame $f_o$ is given by
$f_o = \left(\left(\sqrt{p + q} - \sqrt{p}\right)/2\right)s$
and the width of the inner frame $f_i$ is given by
$f_i = \left(\left(1-\sqrt{p + q}\right)/2\right)s$, where $p$ is defined as the ratio of the inner-most square over the total area, and $q$ is the ratio of the area of the inner frame over the total area.
To obtain a training, validation, and test sets ratio of 2:1:1 similar to BigEarthNet, $p=q=0.25$ are selected\footnote{The \gls{lulc} class distribution is available at: \url{https://bigearth.net/static/documents/BigEarthNet_v2_Split.pdf}}.

\subsection{Supplementary software tools}\label{subsec:software-tools}
The \BENnew{} patches are available in the same format (GeoTIFF) as in the \BEN{} dataset.
However, this format is not optimized for training and evaluating \gls{dl} models.
The training and evaluation time can be greatly reduced by converting the files into a beneficial \gls{dl} format that removes unnecessary metadata and stores only the image data (which is stored in such a way that high random read throughput can be achieved without requiring modifications to the underlying data).
To this end, we provide a data format conversion tool \hl{(rico-hdl\footnotemark{})} that takes the patch and stores the name of the patch as well as its data as a key-value pair in an LMDB \cite{lmdb} database, where the data is encoded in the \gls{dl} library agnostic safetensor \cite{safetensor} format.
Utilizing a \gls{dl} library independent encoding is important to ensure that researchers can continue to use the library of their choice while benefitting from significantly faster data access. 
\hl{To ensure that the conversion into the optimized format remains accessible, we publish rico-hdl as a standalone, self-contained Linux binary 
and provide in-depth documentation about its usage.}

\footnotetext{\url{https://github.com/rsim-tu-berlin/rico-hdl}}

\begin{table*}[bt!]
    \centering
    \sisetup{
    table-format=2.2, 
    round-mode=places,
    round-precision=2, 
    drop-uncertainty=true,
    table-figures-uncertainty=1,
    minimum-integer-digits=2,
    minimum-decimal-digits=2
    }
    \footnotesize
    \caption{Experimental results (in $\%$) of the considered DL models trained on \BENnew{} using:
    i) only the Sentinel-1 images (denoted \\ as S1);
    ii) only the Sentinel-2 images (denoted as S2); and
    iii) both S1 and S2 together (denoted as S1+S2) 
    }
    \label{tab:exp}
\begin{tabularx}{.8\linewidth}{X c SSSS SS}
        \toprule
{Model}&{Modality of \BENnew{}}&{$\text{AP}^\text{M}$}&{$\text{AP}^\mu$}&{$F_1^\text{M}$\hspace{1em}}&{$F_1^\mu$\hspace{1em}} \\ \midrule 
{\multirow{3}{*}{ResNet-50 \cite{resnet}}}&S1&62.68051266670227(0.13935397025145876)&80.64591089884439(0.5006812826575512)&56.89904491106669(0.7343067906644171)&70.68190375963846(0.4336835408065224) \\
{}&S2&70.71622411410013(0.8765738042593232)&85.86130738258362(0.06572343745113543)&64.73899881045023(0.9929290779907131)&76.33978327115378(0.19138926699108125) \\
{}&S1+S2&70.76340516408285(0.38585535827849016)&86.05921864509583(0.12208897288637807)&64.89677429199219(0.6941508297771976)&76.50250196456909(0.14173196498690857) \\ \midrule 
{\multirow{3}{*}{ResNet-101 \cite{resnet}}}&S1&61.765060822168984(0.9928963084992656)&80.14883597691855(0.8500636859417491)&55.77226678530375(2.3602213413073545)&69.96570825576782(1.5263108376142387) \\
{}&S2&70.63056627909342(0.20325890916937095)&85.91510454813638(0.2471487920185619)&64.19101556142172(0.3440522277906186)&76.13365451494852(0.25389205188515074) \\
{}&S1+S2&70.92801133791605(0.2786731497744306)&86.20864550272623(0.23127169249815083)&64.69026009241739(0.3170979367101888)&76.5938917795817(0.14969695040803396) \\ \midrule
{\multirow{3}{*}{MLP-Mixer Base \cite{mlpmixer}}}&S1&55.40969967842102(0.4376252242454917)&74.15493925412497(0.2584305078315487)&49.90881383419037(0.23419344165418263)&65.00853697458902(0.07440103372382195) \\
{}&S2&67.77195334434509(0.13208110568397344)&84.32462016741434(0.1650899754203384)&62.485384941101074(1.0450263411489202)&74.59068099657694(0.15177024089420313) \\
{}&S1+S2&68.21700930595398(0.40436930147665284)&84.65690612792969(0.2706652621129021)&62.74266242980957(0.4925919315750041)&74.82598622639975(0.340112899221138) \\ \midrule
{\multirow{3}{*}{MobileViT S \cite{mobilevit}}}&S1&62.68837253252665(0.4341943787681634)&80.9980551401774(0.16522352919142988)&55.84314068158468(0.8915196316936947)&70.5804189046224(0.08666736084218588) \\
{}&S2&69.83956098556519(0.5566227882645092)&86.20014389355978(0.22557860296128412)&62.10054357846578(0.491288359326901)&75.99181731541952(0.26343845436184155) \\
{}&S1+S2&70.3260064125061(0.2538308449413237)&86.20814681053162(0.02424455029459704)&63.01760276158651(0.6416661758034343)&76.14953319231668(0.11078128359005406) \\ \midrule
{\multirow{3}{*}{\added{MobileNet V4 Hybrid Medium \cite{mobilenetv4}}}}&S1&61.232441663742065(0.23098038524822184)&80.39411902427673(0.03454854477644492)&56.22666080792745(0.5930783532428552)&70.46001354853311(0.12524689220918026) \\
{}&S2&68.68526140848795(0.37533824453767617)&85.45194268226624(0.16319886502478587)&62.470410267512(0.3879148000025014)&75.55299599965414(0.10452132241751638) \\
{}&S1+S2&68.96708607673645(0.08930640784254132)&85.46791474024454(0.09304727994446446)&63.175167640050255(0.4860835332904035)&75.67503452301025(0.1720681168091059) \\ \midrule
{\multirow{3}{*}{ConvNext V2 Base \cite{convnextv2}}}&S1&59.247368574142456(0.6434817494073738)&78.29042871793112(0.06341345142058483)&54.545044898986816(0.3668465427268385)&68.92192761103311(0.3401111363498125) \\
{}&S2&68.60916018486023(0.1625939287528044)&85.13185779253641(0.11553348455212104)&62.639488776524864(0.6119263027786827)&75.42755802472433(0.28566502185625725) \\
{}&S1+S2&68.77015233039856(0.4485255244558742)&85.50865650177002(0.2731434332663591)&62.688408295313515(0.6750898920380032)&75.94518462816873(0.4061443833098033) \\ \midrule
{\multirow{3}{*}{\added{InceptionNeXt Base \cite{inceptionnext}}}}&S1&61.047824223836265(0.359950392121398)&79.71473932266235(0.07739384650397377)&55.901686350504555(0.4664268964565351)&69.89054679870605(0.0883575967123425) \\
{}&S2&69.3224847316742(0.4422249145105657)&85.4844868183136(0.18389930532509943)&62.93149987856547(0.8674579019242852)&75.63303510348001(0.47376882032161577) \\
{}&S1+S2&69.56921815872192(0.22154974759326815)&85.66885193188986(0.12740895997373727)&63.381979862848915(0.9901502484512313)&76.14955306053162(0.29093939035766114) \\ \midrule
{\multirow{3}{*}{\added{RDNet Base \cite{rdnet}}}}&S1&58.53773156801859(0.10023667059383279)&78.038889169693(0.012892104380021375)&52.20238367716471(0.3616983146300344)&67.852383852005(0.31020717058061104) \\
{}&S2&68.53371858596802(0.5388470744560991)&85.41714946428934(0.02841891478613844)&62.34876314798991(0.13864276758372548)&75.6211002667745(0.12374098034038561) \\
{}&S1+S2&68.70301167170206(0.16782546497047093)&85.6906751791636(0.3219565074233191)&62.458970149358116(0.5498546162997278)&75.98844567934671(0.25819067765252823) \\
\bottomrule
\end{tabularx}
\end{table*}

\section{Experimental results}\label{sec:experimental-results}
In the experiments, 
we considered the following state-of-the-art model configurations: i) two ResNets with different numbers of layers (ResNet-50 and ResNet-101) \cite{resnet};
ii) MLP-Mixer Base \cite{mlpmixer};
iii) MobileViT S \cite{mobilevit};
iv) MobileNet V4 Hybrid Medium \cite{mobilenetv4}; 
v) ConvNeXt V2 Base \cite{convnextv2};
vi) InceptionNeXt Base \cite{inceptionnext}; and
vii) RDNet Base \cite{rdnet}.

ConfigILM \cite{configilm} is utilized to train the considered models in the framework of multi-modal \gls{rs} image scene classification.
The associated code\footnotemark{}, as well as the pre-trained weights\footnotemark{} for broader research and application use-cases in \gls{rs}, are publicly available.
\footnotetext[5]{\url{https://git.tu-berlin.de/rsim/reben-training-scripts}}
\footnotetext{\url{https://huggingface.co/BIFOLD-BigEarthNetv2-0}}


In the experiments, we did not use the Sentinel-2 image bands associated with \SI{60}{\meter} spatial resolution (bands 1 and 9).
This is due to the fact that these bands are primarily utilized for cloud screening, atmospheric correction, and cirrus detection in \gls{rs} applications, and do not contain a substantial amount of information for the characterization of the semantic content of \gls{rs} images \cite{BigEarthNetMM}.
The lower resolution \SI{20}{\meter} Sentinel-2 bands were upsampled via nearest neighbour interpolation to the size of the \SI{10}{\meter} resolution bands.
To achieve multi-modal learning, we stacked the Sentinel-1 and the Sentinel-2 bands into one volume.
As recommended in \cite{BigEarthNetMM}, we did not consider the patches that are covered by snow, clouds or cloud shadows.
The train, validation and test sets were obtained by applying the geographical-based split assignment algorithm.
All models were trained three times with different seeds for up to \SI{100}{\epochs} with the AdamW optimizer and a linear-warmup-cosine-annealing learning rate of \num{1e-3} after \num{1000} warm-up steps with batch size, dropout and drop path set to \num{512}, \num{0.15} and \num{0.15}, respectively. The averaged results across the three runs are reported.
All models were trained on a single H100 GPU.
To evaluate the results, we employ two average precision (macro and micro averaged) metrics (denoted as $\text{AP}^\text{M}$/$\text{AP}^{\mu}$) and the macro and micro averaged $F_1$ score ($F_1^\text{M}$/$F_1^{\mu}$).
The results of the considered models obtained when using \BENnew{} with: i) only the Sentinel-1 images (denoted as S1); ii) only the Sentinel-2 images (denoted as S2); and iii) both S1 and S2 together (denoted as S1+S2) are shown in \Cref{tab:exp}.
\added{By assessing the table, one can observe that the use of S2 improves the performance across all metrics compared to that of S1 only.
In addition, the joint use of S1 and S2 results in the highest performance compared to those obtained by using either S1 or S2 only.
As an example, ResNet-101 achieves an $\text{AP}^\text{M}$ performance of \SI{70.93}{\percent} with S1+S2 which is \SI{0.30}{\percent} and \SI{9.16}{\percent} better than trained on S2 and S1, respectively.
In detail, when S1 and S2 are jointly considered, the ResNet models generally lead to the highest performance.
For example, in terms of $\text{AP}^\text{M}$, the ResNet-50 model (second best) achieves \SI{70.76}{\percent}, \SI{0.17}{\percent} less than the  ResNet-101 (best-performing model) and \SI{0.43}{\percent} more than the MobileViT S (third best) model, respectively.
The lowest performance is, in general, obtained by the MLP-Mixer Base model. As an example, the use of the MLP-Mixer model yields an $\text{AP}^\text{M}$ of \SI{68.22}{\percent}, which is more than \SI{2.7}{\percent} lower than that obtained by ResNet-101.}

\section{Conclusion and Discussion}\label{sec:conclusion-and-discussion}
In this paper, we have introduced refined BigEarthNet (\BENnew{}) that significantly improves the BigEarthNet dataset by:
1) utilizing the most recent
version (2.11) of the sen2cor atmospheric correction tool 
and removing the tiles with failing quality indicators from the dataset;
2) overcoming the \gls{lulc} label noise present in BigEarthNet by utilizing the latest CLC map to derive the labels, while also making the pixel-level reference maps available;
3) proposing a new recommended split that utilizes a geographical-based split assignment algorithm to reduce the spatial correlation among the training, validation, and test sets \hl{by assigning the geographically overlapping patches of the same season only to the training set};
4) providing new software tools to efficiently train DL models; as well as 5) releasing pre-trained weights obtained by using state-of-the-art DL models.
We expect that \BENnew{} will lead to more reliable and interpretable research results for image classification and retrieval in \gls{rs} compared to the \BEN{} dataset.

\pagebreak
\small
\balance
\printbibliography[title=References]

\end{document}